%% file: main.tex
\documentclass[10pt,letterpaper]{article}

\usepackage{cogsci}

\input{math_commands}

\cogscifinalcopy 

\usepackage[font=small,skip=0pt]{caption}
\usepackage{pslatex}
\usepackage{apacite}
\usepackage{graphicx}
\usepackage{float} 
\usepackage{subcaption}
\usepackage[compact]{titlesec}
\titlespacing*{\section}{1pt}{*1}{*1}

\usepackage{graphicx,amsmath,amsfonts,amssymb}
\usepackage{xcolor}         
\usepackage{comment}
\usepackage{blindtext}
\usepackage[colorinlistoftodos,prependcaption,textsize=tiny]{todonotes}
\usepackage{regexpatch}
\usepackage{multicol}
\usepackage{multirow}
\usepackage[font=small]{caption}
\usepackage{float}
\usepackage{graphicx}
\usepackage{nicefrac}       
\usepackage{microtype}      
\usepackage{booktabs}       
\usepackage[font=small]{caption}
\usepackage{float}

\usepackage{wrapfig}

\definecolor{cadmiumgreen}{rgb}{0.0, 0.42, 0.24}
\definecolor{ao(english)}{rgb}{0.0, 0.5, 0.0}

\usepackage{url}

\title{Improving Systematic Generalization Through Modularity and Augmentation}

\author{{\large \bf Laura Ruis (laura.ruis.21@ucl.ac.uk)} \\
  University College London
  \AND {\large \bf Brenden Lake (brenden@nyu.edu)} \\
  New York University; Facebook AI Research}

\begin{document}
\setlength{\abovedisplayskip}{0pt}
\setlength{\belowdisplayskip}{0pt}
\setlength{\abovedisplayshortskip}{0pt}
\setlength{\belowdisplayshortskip}{0pt}

\maketitle

\begin{abstract}
Systematic generalization is the ability to combine known parts into novel meaning; an important aspect of efficient human learning, but a weakness of neural network learning. In this work, we investigate how two well-known modeling principles---modularity and data augmentation---affect systematic generalization of neural networks in grounded language learning. We analyze how large the vocabulary needs to be to achieve systematic generalization and how similar the augmented data needs to be to the problem at hand. Our findings show that even in the controlled setting of a synthetic benchmark, achieving systematic generalization remains very difficult. After training on an augmented dataset with almost \emph{forty times} more adverbs than the original problem, a non-modular baseline is not able to systematically generalize to a novel combination of a known verb and adverb. When separating the task into cognitive processes like perception and navigation, a modular neural network is able to utilize the augmented data and generalize more systematically, achieving 70\% and 40\% exact match increase over state-of-the-art on two gSCAN tests that have not previously been improved. We hope that this work gives insight into the drivers of systematic generalization, and what we still need to improve for neural networks to learn more like humans do.

\textbf{Keywords:} 
Modularity; Systematic Generalization; Data Augmentation
\end{abstract}

\input{main_text_cogsci}



\bibliographystyle{apacite}

\setlength{\bibleftmargin}{.125in}
\setlength{\bibindent}{-\bibleftmargin}

\bibliography{CogSci_Template}

\end{document}

%% file: math_commands.tex
\usepackage{amsmath,amsfonts,bm}

\def\eqref#1{equation~\ref{#1}}

\def\1{\bm{1}}

\DeclareMathAlphabet{\mathsfit}{\encodingdefault}{\sfdefault}{m}{sl}
\SetMathAlphabet{\mathsfit}{bold}{\encodingdefault}{\sfdefault}{bx}{n}



%% file: main_text_cogsci.tex
\section{Introduction}

Humans are efficient learners. Once someone has seen a single example of a wampimuk, they know how to recognize a small wampimuk. Once someone learns how to walk cautiously, they know how to cycle cautiously, even though cycling requires a novel sequence of low-level actions. Our aptitude for this type of generalization is a consequence of the fact that word meanings like ``small" and ``cautiously" compose systematically \cite{Chomsky:1957,Montague:1970a} --- a property of language called systematic compositionality. 

A discrepancy between humans' ability to interpret novel compositions and that of neural network models has long been discussed \cite{Fodor:Pylyshyn:1988,Marcus:1998,Fodor:Lepore:2002,Marcus:2003,Calvo:Symons:2014} and this issue has been revitalized in recent years \cite{Gershman:2015,Lake:Baroni:2017,Bastings:etal:2018,Loula:etal:2018,bahdanau:etal:2018}. While conventional models assign low likelihood to unseen combinations of familiar tokens, there has been a recent surge of interest in developing models that generalize more systematically \cite{Russin:2019,lake:2019,andreas:2019,nye:2020,Gordon:2020,Bogin:2021}. Despite progress, we are still very far from neurally-grounded models that provide a fully satisfying account of systematic compositionality, and certain types of complex composition remain especially elusive.

The methods that systematically handle prototypical examples of compositionality like verb-object binding do not address other less studied cases like adverb-verb composition. Similarly to the ``cycle cautiously"-example above, this type of generalization requires making non-local changes to the action sequence, transforming it entirely. The example reflects a more general property of adverbs; combining a verb with an adverb in a sentence can substantially change the actions required to perform said verb. We study this phenomenon using the recently proposed gSCAN benchmark for evaluating models for systematic reasoning in a controlled environment \cite{Ruis:etal:2020}. The grounded nature of this benchmark allows it to evaluate new dimensions of systematicity that have not been captured by previous work, like the additional complexity of adverb-verb compositionality. Grounding meaning in a world state requires models to do something that is more like real language understanding; the correct action sequence for a language command changes based on the world state. Several recent works have proposed methods to improve performance on the tests \cite{Heinze:Bouchacourt:2020,gao:et:al:2020,kuo:Katz:barbu:2021,nye:tessler:tenenbaum:lake:2021,qiu:et:al:2021,Jiang:Bansal:2021}, yet there has been little progress on the types of unseen adverb combinations discussed above.

Here, we study two key modeling principles, modularity and structured data augmentation, and their effect on systematic generalization in grounded language learning. The notion that modular architectures generalize better is a longstanding principle in cognitive science \cite{Fodor:1983} and AI \cite{poole:mackworth:2017}, just as programmers know that modular systems are more reusable and robust \cite{meyer:1997}. Densely-connected neural networks may be especially prone to the pitfalls of non-modular systems, as spurious correlations in the input can affect the whole system in unpredictable ways. There is ample evidence that modularity can improve systematic generalization in neural-network-style models,  \cite{andreas:etal:2016,bahdanau:etal:2018,purushwalkam:2019,damario:etal:2021} and in this work we apply modularity to the gSCAN challenge, decoupling navigation, perception, and reasoning.


Data augmentation is a widely used technique in machine learning used to increase the diversity of training examples without explicitly collecting new data (for surveys in computer vision and natural language processing see \citeauthor{Shorten:2019} \citeyear{Shorten:2019} and \citeauthor{feng:etal:2021} \citeyear{feng:etal:2021} respectively). However, applying data augmentation techniques to get examples for adverb-verb composition is difficult. The transformative effect an adverb has on the output sequence and the grounded nature of the benchmark makes the class of augmentation techniques called interpolation-based, like those proposed by \citeauthor{andreas:2019} \citeyear{andreas:2019} and \citeauthor{kagitha:2020} \citeyear{kagitha:2020}, inapplicable. Instead we take a structured augmentation approach and infer the set of rules that govern adverbs, subsequently using them to generate new data. We evaluate how much and what kind of experience is needed to generalize to examples exhibiting the type of compositionality illustrated by the ``cycle cautiously''-example.

In this paper we propose a modular approach to systematic generalization and design a structured data augmentation technique to generate experience for the modules. From extensive experiments with the proposed setup we can deduce three main findings. Firstly, simply providing a neural network with more data might not be sufficient to achieve systematic generalization; even after adding as much as 150 extra adverbs to the four original adverbs in the gSCAN dataset (arguably enough given the simple, controlled environment of the tests) the model is not able to generalize systematically. Secondly, implementing modularity by separating the problem into high-level cognitive processes enables the model to utilize the additional experience and generalize systematically on tests related to the additional experience. Finally, naively adding experience does not enable systematic generalization. For the model to learn how to generalize systematically, the experience it sees needs to be sufficiently similar to the type of systematicity it is tested on. 

The three key contributions of this paper are (1) We propose a neural architecture for systematic generalization that is modular at higher levels of cognition, (2) We analyse the amount and type of data this network needs to achieve systematic generalization, and (3) We improve the SOTA on two tests of the recently proposed gSCAN benchmark.

\section{Related Work}

In recent years systematic generalization has seen a revived interest from the machine learning community, in computer vision \cite{Johnson:2017,misra:etal:2017,atzmon:2020,ruis:etal:2021}, natural language processing \cite{Lake:Baroni:2017,baroni:2019,keysers:2020,kim:linzen:2020}, and more generally \cite{nam:mcclelland:2021}. Two fundamentally different approaches are taken by the literature; one utilizes additional data while making few changes to the conventional setup and architecture \cite{furrer:etal:2020}, while the other utilizes additional inductive biases that aim to support systematic generalization \cite{Russin:2019,lake:2019,andreas:2019,nye:2020,Gordon:2020,Bogin:2021,chaabouni:etal:2021}. In this work we apply both approaches, the former through data augmentation, and the latter through high-level modularity.

Using modularity to improve systematic generalization is a common strategy in the literature. Some approaches introduce modules that directly map to different parts of the input command \cite{andreas:etal:2016,corona:etal:2021}, and others map to different attributes of an input image \cite{purushwalkam:2019}. In this work we design modules that map to different high-level cognitive processes, like navigation and perception. Additionally, we use data augmentation to provide the modules with separate experience. Data augmentation for systematic generalization is also a well-studied area \cite{lake:2019,andreas:2019,kagitha:2020}. Here we take a structured augmentation approach that works even in the grounded setting of the benchmark, designing a set of rules that can generate novel examples. This allows us to generate separate data for each module as well as additional data with novel concepts that are not part of the original dataset.

\section{Evaluating Systematic Compositionality}

\begin{figure*}[ht]
\vspace*{-\baselineskip}
\begin{subfigure}{.45\textwidth}
  \centering
  \centerline{\includegraphics[scale=0.18]{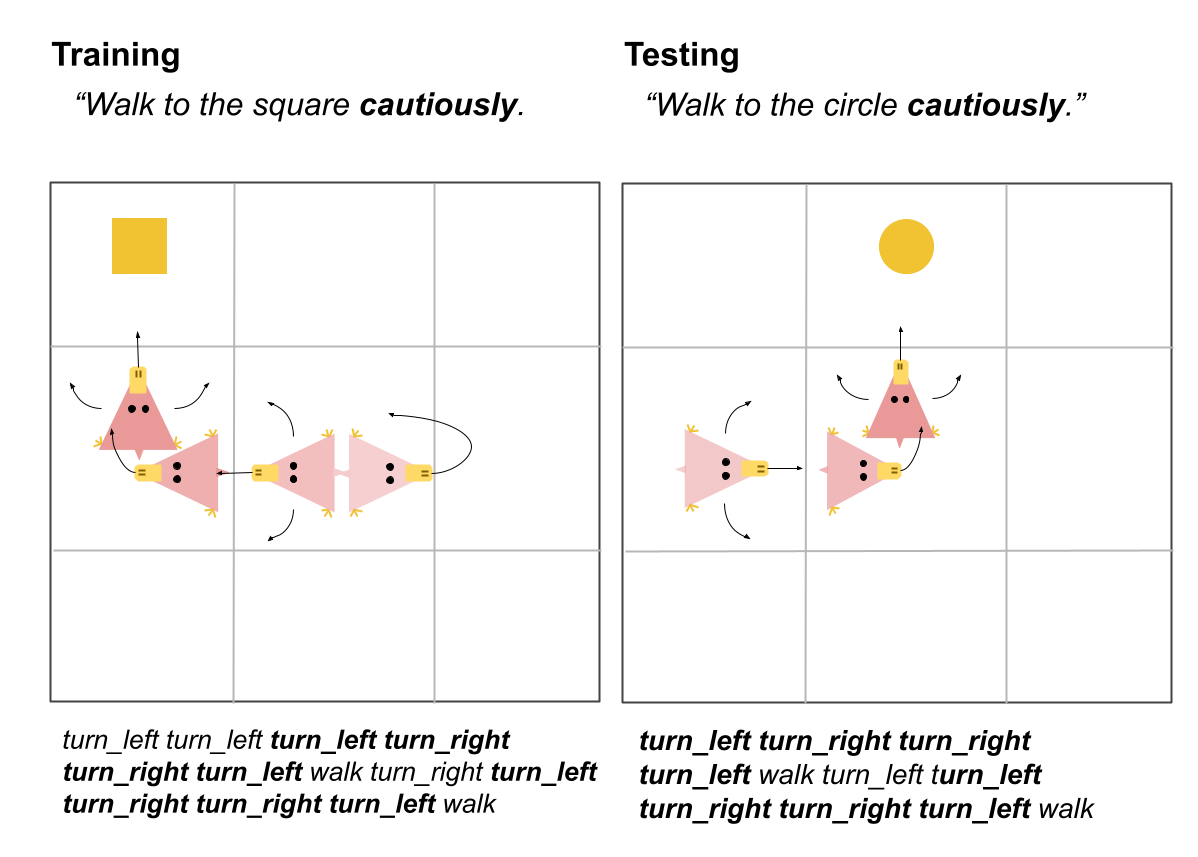}}
  \caption{}
  \label{fig:sub-first}
\end{subfigure}\hfill%
\begin{subfigure}{.45\textwidth}
  \centering
  \centerline{\includegraphics[scale=0.18]{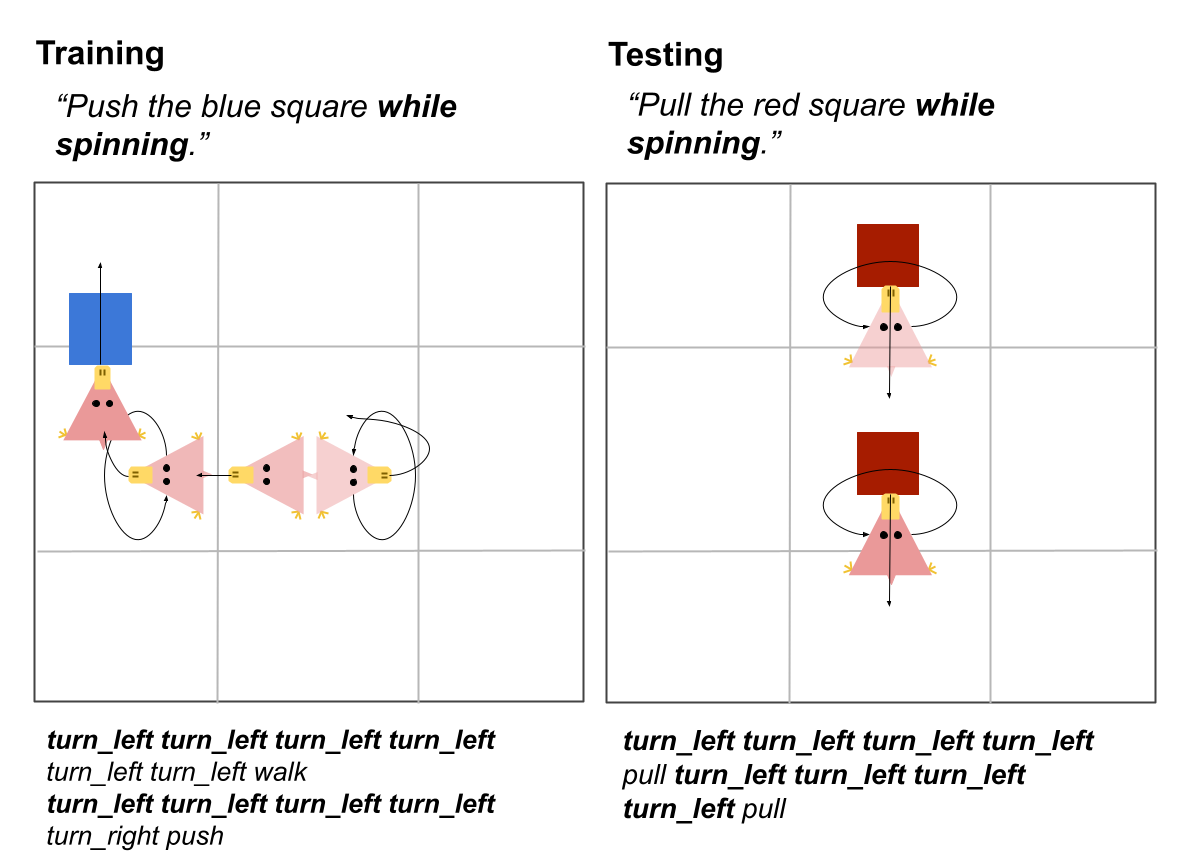}}
  \caption{}
  \label{fig:sub-second}
\end{subfigure}
\caption{This figure depicts the two tests of adverb compositionality in gSCAN. Figure (a) denotes a few-shot learning test; a model has access to few (\textit{k}) examples of how the adverb ``cautiously'' translates to an output sequence and needs to generalize to all other examples. Figure (b) denotes the ``pull while spinning''-test; reminiscent of the ``cycle cautiously''-example, a model learns all examples of pushing while spinning or walking while spinning, and is tested on its ability to interpret ``pull while spinning''.} 
\label{fig:adverb-split-exs}
\vspace*{-\baselineskip}
\end{figure*}

To evaluate the systematicity of the model's predictions, we will use the grounded SCAN benchmark (gSCAN). The benchmark tests a broad set of phenomena in situated language understanding where humans should easily generalize, but where computational models struggle due to the systematicity of the differences between the training data and the test data. In gSCAN, agents are asked through language commands to execute instructions in a 2D gridworld. The benchmark has seven different tests that require combining known concepts into novel meaning. For example, one test requires recognizing a novel object with a familiar color and shape that have never been observed together. Another test evaluates whether a model is able to interpret a command containing an unseen composition of a familiar verb and adverb. This latter test is of the kind that is unsolved by prior work, hence the subject of our focus.

Figure \ref{fig:adverb-split-exs} depicts examples of gSCAN tasks with adverbs. The use of an adverb in the input command indicates a changed manner of navigation across the grid, requiring a complete transformation of the output sequence. For example, transforming a sequence that simply navigates from the bottom right to the top left to one that does this while spinning works as follows: 
\begin{align*}
    &\text{{\small``turn\_left walk walk turn\_left walk walk"}} \\
    & \quad \quad \downarrow_{f_\text{while spinning}} \\
    &\text{{\small``spin* turn\_left walk spin* walk spin* turn\_left walk spin* walk"}}
\end{align*}
where spin* is not a primitive but an action sequence of four times ``turn\_left". The underlying rule is to add a spin before turning to a direction and moving (e.g.,  ``turn\_left walk'' or ``turn\_right push''). When the manner is ``cautiously'', the transformed sequence would look as follows:
``turn\_left cautious* walk cautious* walk turn\_left cautious* walk cautious* walk'', where cautious* is an action sequence of ``turn\_left turn\_right turn\_right turn\_left'' (looking to the left and right). Here the rule is slightly different, as the agent first needs to turn to the direction it will continue in, then be cautious, and then take the movement action. This difference will prove important in the experiment section where we analyze what kind of data augmentation results in systematicity.

Figure \ref{fig:adverb-split-exs} exemplifies the two types of systematic generalization of our focus and that have not been solved by prior work. Figure \ref{fig:sub-first} shows a test of few-shot generalization to ``cautiously.'' At training time, the model only sees a few ($k$) examples of commands with ``cautiously''; at test time, the model needs interpret all unseen commands with that manner. Figure \ref{fig:sub-second} shows a test of adverb-to-verb generalization. At training time, the model sees all commands that contain ``push ... while spinning'' and ``walk ... while spinning'', and at test time it needs to interpret ``pull ... while spinning''.

\section{Method} \label{method}
The problem posed by gSCAN naturally divides into modules of high-level cognitive processes. Below we describe what each module does, how we generate data for these modules, and how the resulting model is trained.

\begin{figure}[H]
\begin{center}
\centerline{\includegraphics[width=0.9\columnwidth]{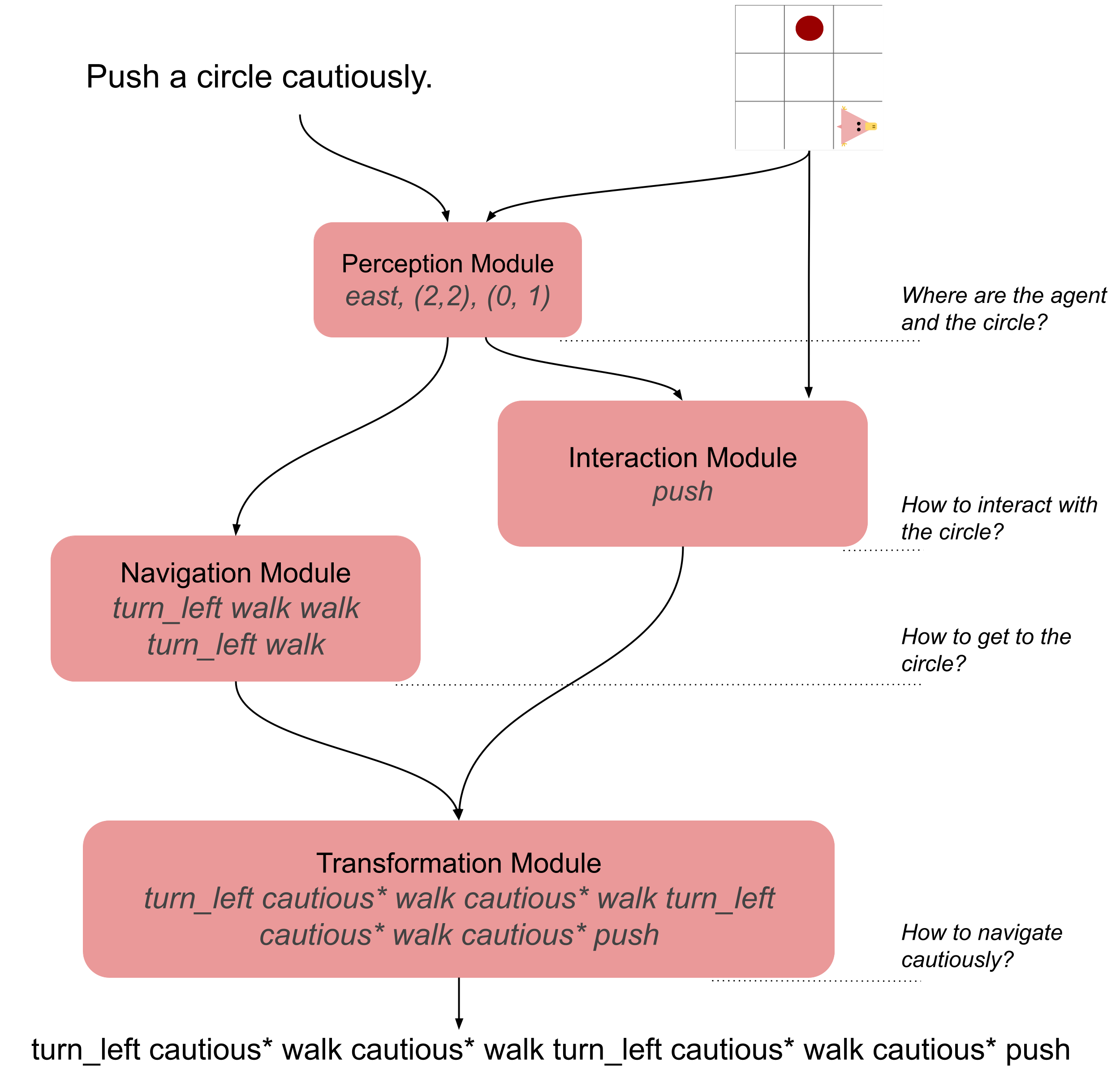}}
\end{center}
\caption{The input command (``Push a circle cautiously.'') and world state are processed by different modules, each dealing with a different question about the input task. The final output is produced by the transformation module. \small{*: cautious is in reality not a primitive action but a sequence of “turn\_left turn\_right turn\_right turn\_left”}} 
\label{fig:modules-layout}
\end{figure}

\textbf{The modular architecture.} To facilitate modularity, we aim to separate the task at hand into different cognitive modules; perception, navigation, interaction, and manner of navigation (or, transformation). Each module has its own input and output, all modules additionally have access to the language command, and some have access to the world state. For an overview of the modules and their input-output flow, see Figure \ref{fig:modules-layout}. 

\textit{The perception module} takes the input command and world state and outputs where the agent is (a tuple with a row and column number), what direction it is facing (east, north, south, or west), and where the target object is (a tuple with a row and column number). \textit{The navigation module} gets all that information and outputs a plan to get from the agent location to the target location. This module needs access to the input command because there are adverbs like ``while zigzagging'' that require the module to output a different plan (instead of ``turn\_left walk walk turn\_left walk'' like the plan in the figure, it would output ``turn\_left walk turn\_left walk turn\_right walk''). Additionally, some adverbs require an egocentric plan like the one in Figure \ref{fig:modules-layout}, but others require an allocentric plan. The allocentric equivalent to the plan in Figure \ref{fig:modules-layout} is ``North North West''. The navigation module decides what mode of output, allo- or egocentric, to use based on the input command. In the next paragraph the reason for this distinction is detailed. \textit{The interaction module} gets the output from the perception module and the world state, and can use the information of the target location to decide how to interact with the object. In this case, because an object always needs to be pushed to the wall, it needs to be pushed once. This module uses its access to the input command to decide whether to push or pull. Finally, \textit{the transformation module} gets the plan and the interaction commands, and transforms the whole sequence to adhere to the manner of navigation. In this case, it needs to navigate cautiously (``turn\_left turn\_right turn\_right turn\_left'', i.e., looking to the left and right, shortened in Figure \ref{fig:modules-layout} for clarity) before each movement command (``walk'', ``push'', and ``pull''). This module looks at the adverb in the input command to decide which transformation is necessary and always outputs egocentric commands, even if the input is allocentric. Inputs like ``North'' are transformed to ``spin* turn\_left walk'' if the adverb is ``while spinning'' (where spin* is ``turn\_left turn\_left turn\_left turn\_left''). Sometimes an example requires no interactions and/or no transformation (e.g., ``walk to the circle''). In that case, the interaction and transformation modules output a special end-of-sequence token.

\textbf{Structured data augmentation.} Each module needs experience to learn, and in order to analyze how much experience they need, we construct a domain-specific language (DSL) under which the original gSCAN benchmark is grammatical. The DSL contains everything to generate the already existing manners of navigation found in gSCAN and can be used to sample new manners of navigation. We can use it to generate as much data as needed. The DSL is constructed as a functional L-system \cite{Lindenmayer:1968}. L-systems are parallel rewriting systems of rules. An L-system has a set of symbols that can form sequences, and production rules to rewrite those sequences. By applying rules from the L-system to sequences (which can be empty), a sequence can be grown. L-systems are executed in parallel, meaning that a rule gets applied to all symbols in the sequence at once. Each adverb gets assigned a set of rewrite rules, called a program. Applying this set of rules to a sequence will transform it into a sequence with the manner that the program corresponds to.

To show how the DSL works, we will walk through an example. The DSL uses allocentric and egocentric symbols, which are indicated by capitalized and lowercase symbols respectively (e.g., ``North'' or ``turn\_left'' and ``walk''). This distinction more naturally fits the already existing adverbs in gSCAN and results in less rewriting rules for their programs. For the adverb ``while spinning'' the transformation can be applied to allocentric commands, and for the adverb ``cautiously'' to egocentric commands. For example, to construct the sequence in Figure \ref{fig:modules-layout}, we apply the following rewrite-rule to the egocentric plan ``turn\_left walk walk turn\_left walk'':
\begin{align*}
    \text{{\small walk}} &\rightarrow \text{{\small turn\_left turn\_right turn\_right turn\_left walk}}
\end{align*}
Which puts the ``cautious''-sequence at the right location, right before the movement symbol.  Because L-systems are parallel rewriting systems, the above rule needs to be applied once to the sequence for it to be transformed. For ``while spinning'', and an allocentric plan ``North North West'', the rewrite rules will be as follows:
\begin{align*}
    \text{{\small North}} &\rightarrow \text{{\small turn\_left turn\_left turn\_left turn\_left North}} \\
    \text{{\small West}} &\rightarrow \text{{\small turn\_left turn\_left turn\_left turn\_left West}}
\end{align*}
If then, based on the agent's starting direction, North gets rewritten to ``turn\_left walk'' the spin is already at the right location in the sequence (namely before turning into the direction that the agent will be walking in). Rules can be applied recursively, which is why the DSL can theoretically generate infinite data. If the above rules are applied ad infinitum the agent will never stop spinning.

Constructing the rules for each adverb leads to a set that we call the meta grammar. We extend the meta grammar with additional rules to be able to sample new programs that correspond to different manners and adverbs. For example, one rule that can generate manners of navigation that aren't a part of the original gSCAN is the following:
\begin{align*}
    \text{{\small East}} &\rightarrow \text{{\small North East South}}
\end{align*}
Applying this type of rule will make the agent take a small detour instead of going east immediately. To get a dataset with $X$ extra adverbs, we sample as many programs from the meta grammar. If a program gets generated that has the same set of rules as one of the original gSCAN adverbs, it is rejected\footnote{The full DSL and the system of rules for each adverb can be found in a public repository: \mbox{\url{https://github.com/ModularCogSci2022/msa}}.}.

The subtle differences among the original gSCAN manners induce a categorization of adverb types; ``while spinning"-type adverbs can be described by transformations that are naturally applied to allocentric commands and do not result in changed movement across grid-cells (they only result in changed movement \emph{within} grid-cells), ``cautiously''-type adverbs are described by transformations that are applied to egocentric commands and also do not result in changed movement, and ``while zigzagging''-type adverbs are applied to allocentric sequences and do result in a changed path across the grid. Finally, the manners resulting from rules that make the agent take a detour are not part of the original gSCAN dataset and are different in that they cause the agent to walk more than needed. We use this taxonomy to evaluate the type of experience a model needs to generalize systematically.

\textbf{Network architecture \& training details}. The neural network we use as a non-modular baseline is exactly the same as the one described by \citeauthor{Ruis:etal:2020} \citeyear{Ruis:etal:2020}, as well as the hyperparameters used. For each module in the modular network we re-use the parts of the baseline architecture that apply. Based on the input-type and output-type a module requires, different neural networks are combined into a module. For example, the perception module's encoder is exactly the same as the baseline, since it also needs to process the input command and the world state. The decoder does not need to output a sequence but three integer predictions (initial agent direction and position, and target position), therefore the decoder is an multi-layer perceptron (MLP) with three output heads. The navigation module processes the input command and the integers produced by the perception module and outputs a sequence, and here the encoder is an MLP and the decoder a recurrent neural network. The full architecture and code for the experiments can be found in the same repository as the DSL. The modules get trained independently and in parallel with ground-truth inputs and targets generated by the DSL. At test time, the forward-pass through the modules happens sequentially and the modules that take the output from the previous module take their predicted output. We train everything with Adam optimizer \cite{adam}. For both the baseline and the transformation module we found that when adding additional augmented data, the neural network size needs to be increased to prevent underfitting. Whenever we add augmented data to a model, we increase the recurrent hidden sizes from 100 to 400, the embedding dimension from 25 to 50, and decrease dropout from 0.3 to 0.2.

\section{Experiments}

To analyze the effect of modularity and data augmentation on systematic generalization we evaluate the models on gSCAN. Besides the two tests of our focus that require the type of adverb compositionality that prior work struggles with, we report the performance on the five remaining tests to ensure that a performance increase doesn't present a trade-off between tests, as well as on a ``random'' test set that is sampled from the same distribution as the training set. This is a sanity check that all models are able to solve the gSCAN task when the test set has no systematic differences with the training set. We compare the proposed models to the end-to-end neural network used in the original gSCAN paper. Upon establishing the best performing model, we run several experiments to understand the performance increase. Specifically, we look at the effect of the number of adverbs in the training set, the type of adverbs in the training set, and the number of ``cautiously''-examples in the training set (i.e., the \textit{k} in k-shot).

\begin{table*}[t]
\centering
{\scriptsize
\caption{Results obtained by the models for each split, showing exact match accuracy (average of 5 runs $\pm$ std. dev.). The rightmost three columns show current state-of-the-art (SOTA) models.}
\label{tab:mainexperiments}
\begin{tabular}{l|c|c|c|c||c|c|c|}
\cline{2-8}
               & \multicolumn{7}{|c|}{\textbf{Exact Match (\%)}}                                         \\ \hline
\multicolumn{1}{|l|}{\textbf{Split}} & \textbf{Baseline} & \textbf{Modular \& Augmentation} & \textbf{Modular only} & \textbf{Augmentation only} & \textbf{\citeauthor{qiu:et:al:2021}} & \textbf{\citeauthor{kuo:Katz:barbu:2021}} & \textbf{\citeauthor{gao:et:al:2020}} \\  \hline  \hline
\multicolumn{1}{|l|}{Random}              & 97.15 $\pm$ 0.46 & 96.34 $\pm$ 0.28  & 96.35 $\pm$ 0.29          & 96.13 $\pm$ 0.28 & \textbf{99.95 $\pm$ 0.02} & 97.32 & 98.6 $\pm$ 0.95  \\  \hline  \hline
\multicolumn{1}{|l|}{Cautiously k=5-shot}          & 1.12 $\pm$ 0.46 &     \textbf{80.04 $\pm$ 6.06} & 11.40 $\pm$ 5.59  &     76.26  $\pm$ 17.18    & - & 10.31  &   -    \\  \hline
\multicolumn{1}{|l|}{Pull while spinning}    & 19.04 $\pm$ 4.08 &  \textbf{76.84 $\pm$ 26.94}   &  25.87 $\pm$ 32.09  &      25.27 $\pm$ 4.89   & 22.16 $\pm$ 0.01 & 21.95 & 33.6 $\pm$ 20.81       \\  \hline \hline
\multicolumn{1}{|l|}{Yellow squares}     & 30.05 $\pm$ 26.76 &  59.66 $\pm$ 23.76 & 59.66 $\pm$ 23.76  &      22.89 $\pm$ 22.40   & \textbf{99.90 $\pm$ 0.06}   & 95.35  &    99.08 $\pm$ 0.69   \\  \hline
\multicolumn{1}{|l|}{Red squares}    & 29.79 $\pm$ 17.70 & 32.09 $\pm$ 9.79 & 32.09 $\pm$ 9.79 &     10.27 $\pm$ 4.09    & \textbf{99.25 $\pm$ 0.91}   & 80.16 &   80.31 $\pm$ 24.51     \\  \hline
\multicolumn{1}{|l|}{Novel direction}   & 0.0 $\pm$ 0.0 & 0.0 $\pm$ 0.0 & 0.0 $\pm$ 0.0 &     0.0 $\pm$ 0.0             & 0.0 $\pm$ 0.0 & \textbf{5.73} & 0.16 $\pm$ 0.12 \\  \hline
\multicolumn{1}{|l|}{Relativity}  & 37.25 $\pm$ 2.85 & 49.34 $\pm$ 11.60 &  49.34 $\pm$ 11.60 &     31.42 $\pm$ 3.91       & \textbf{99.02 $\pm$ 1.16} & 75.19 &   87.32 $\pm$ 27.38    \\  \hline
\multicolumn{1}{|l|}{Class inference}   & 94.97  $\pm$   1.12 & 94.16 $\pm$ 1.25  & 94.06 $\pm$ 1.21  &  93.65    $\pm$    2.49   & \textbf{99.98 $\pm$ 0.01}  & 98.63 & 99.33 $\pm$ 0.46   \\  \hline
\end{tabular}}
\end{table*}

\textbf{The effect of modularity and augmentation.} We train the baseline and the modular model outlined above on the original gSCAN training set containing four adverbs and on a training set containing 150 additional adverbs generated by the DSL. In Table \ref{tab:mainexperiments} the results on a random test set with no systematic differences with the training set and on the seven systematic generalization tests is shown. When looking at the tests of our focus, row ``Cautiously k=5-shot'' and ``Pull while spinning'', we see that the modular network (column ``Modular only") gives a clear improvement over the non-modular baseline (column ``Baseline") for the few-shot learning test. For the adverb-to-verb generalization there is no clear improvement from modularity alone. However, once we add the augmented dataset with 150 extra adverbs the performance jumps significantly (column ``Modular \& Augmentation''). This provides an improvement of almost 80\% exact match over the baseline for the few-shot learning split, and 55\% for the ``pull while spinning"-split.

The question is whether modularity is necessary; what if we simply train the baseline with the augmented data? This experiment is depicted in the column labeled ``Augmentation only'' of Table \ref{tab:mainexperiments}. We observe two things in this result. Firstly, the modular method with augmentation outperforms the baseline with augmentation for the ``pull while spinning''-split. Secondly and more importantly, training the non-modular baseline with the augmented dataset results in worse performance than the baseline for four of the five other tests in gSCAN. One interesting possible explanation for this is the fact that the augmented data has the same systematic differences with the test sets as the original training set and more exposure to this data increases confidence in the biases extracted from the original training set. For example, in the augmented data generated yellow squares are still never mentioned in the input command. This means that the performance increase for the adverb-splits becomes a trade-off, whereas in the modular case we can simply only provide the transformation module with the augmented data to circumvent this issue.

The best method from literature for the 5-shot learning split achieves 10.31\% exact match \cite{kuo:Katz:barbu:2021}, and for the pull while spinning split the best method achieves 33.6\% exact match \cite{gao:et:al:2020}, meaning this work is the first to make meaningful progress on these tests. Even though our method is advantaged through the augmented adverb set, an important result is evident: to make progress on systematic generalization simply adding extra data is not always enough. In this case, we use high-level modularity to make full use of the structured data augmentation.

\begin{figure*}[ht]
\vspace*{-\baselineskip}
\begin{subfigure}{0.3\textwidth}
  \centering
  \centerline{\includegraphics[scale=0.35]{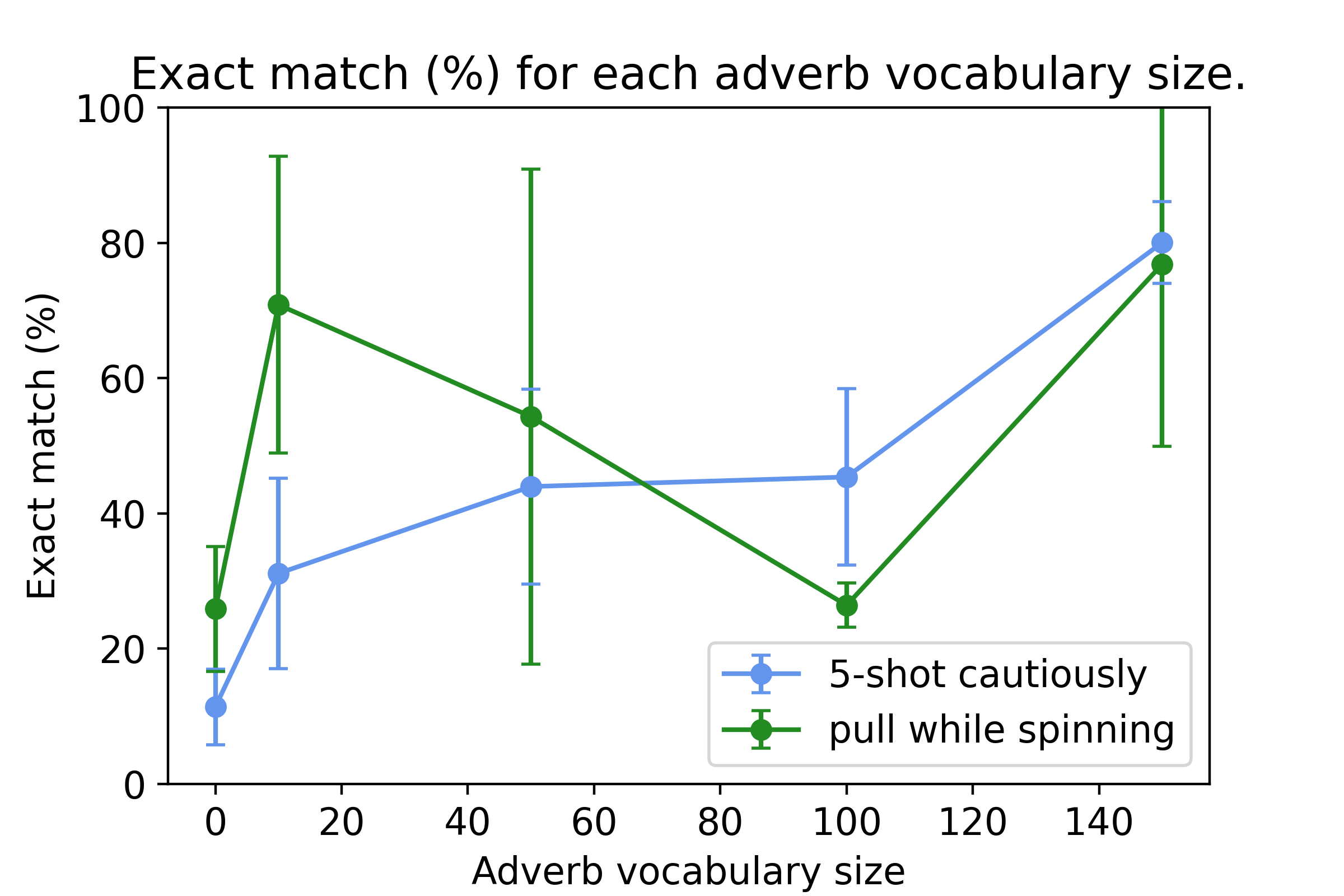}}
  \caption{}
  \label{fig:vocab-size}
\end{subfigure}\hfill%
\begin{subfigure}{0.3\textwidth}
  \centering
  \centerline{\includegraphics[scale=0.35]{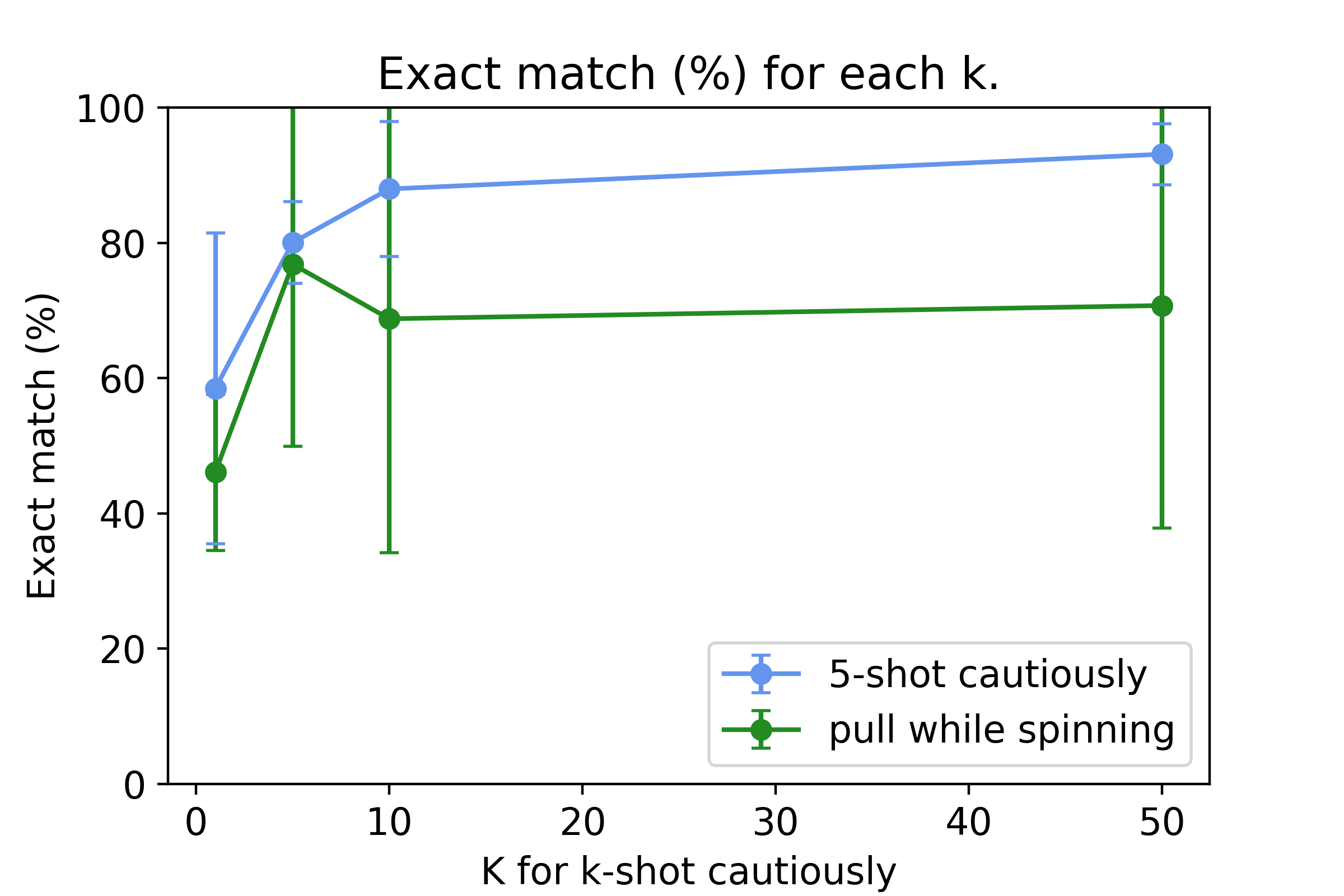}}
  \caption{}
  \label{fig:k}
\end{subfigure}\hfill%
\begin{subfigure}{0.3\textwidth}
  \centering
  \centerline{\includegraphics[scale=0.35]{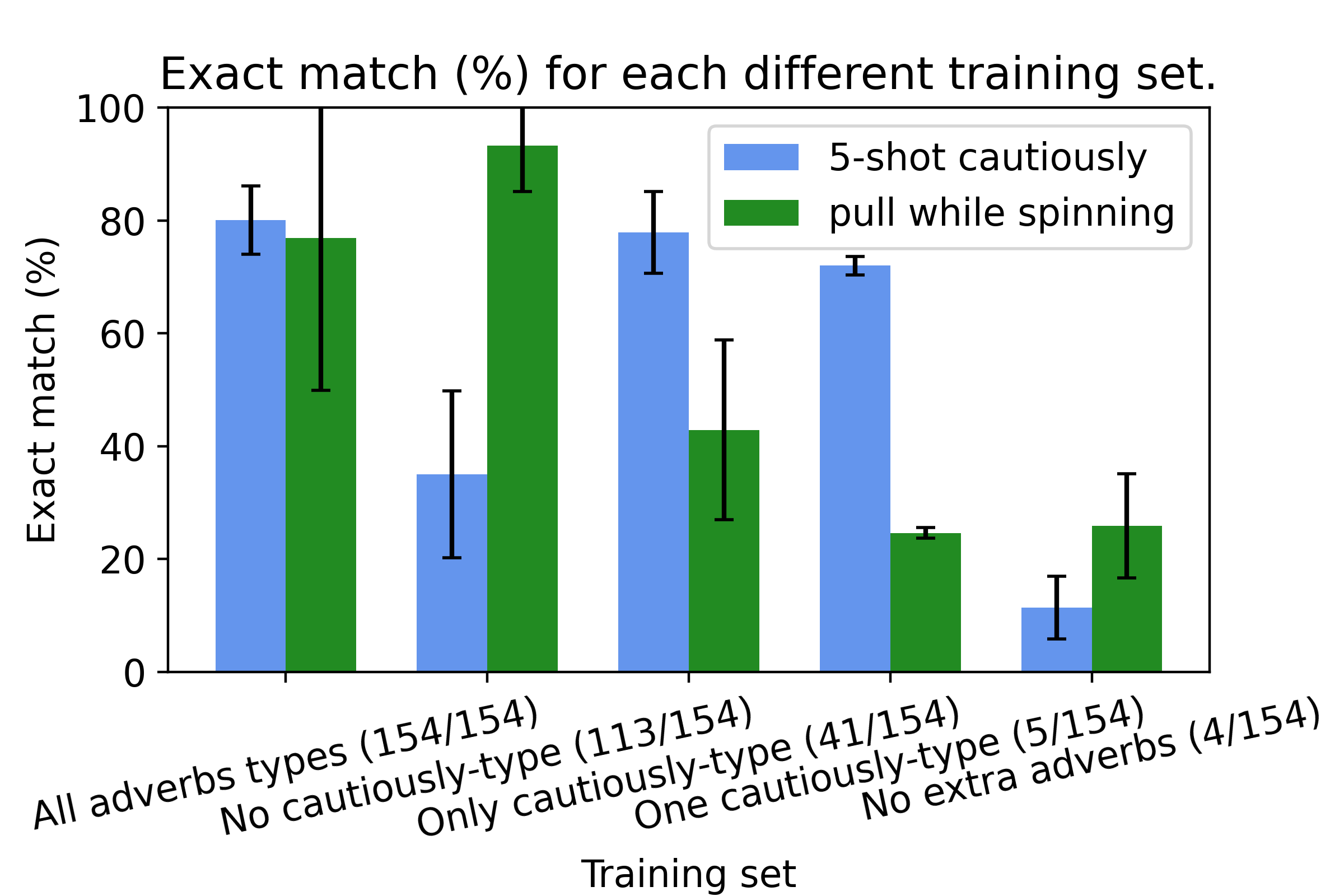}}
  \caption{}
  \label{fig:type}
\end{subfigure}
\caption{The average exact match of five runs on the adverb splits varying (a) the number of additional adverbs (0, 10, 50, 100, 150), (b) the k in k-shot learning (1, 5, 10, 50), or (c) the distribution of adverb types in the training set. The error bars show the std. dev. over 5 training runs.}
\label{fig:vocab-size-k}
\vspace*{-\baselineskip}
\end{figure*}

\textbf{The effect of vocabulary size and varied k.} Figure \ref{fig:vocab-size} contains the results of runs on training sets with different vocabulary sizes. For the 5-shot ``cautiously" split, what we observe is a slow increase of performance when going from 0 to 100 adverbs in the training set, and a big jump when adding the final 50 adverbs. For the ``pull while spinning" split adding extra adverbs is not always strictly better\footnote{In the training sets containing 0, 10, 50, or 150 adverbs the dominant adverb type is the ``while spinning''-type. In the training set with 100 extra adverbs however, the dominant adverb-type is the ``cautiously''-type (3\% more ``cautiously''-type examples). This might explain the performance dip for the ``pull while spinning'' test when training on 100 adverbs. To properly test this hypothesis more experiments are needed.}. This result shows that simply adding more adverbs does not guarantee performance increase. If the model could transfer experience with any adverb to the two that are the subject of the adverb tests we should see a steady increase in performance when adding additional adverbs. What we can infer from these results is that the vocabulary size cannot be the full picture, and we need to break the dataset down per adverb type to get a clearer idea, which we do in the next section. 

Figure \ref{fig:k} shows the performance as a function of the number of examples of ``cautiously'' in the training set. The result is unsurprising; more is better, and after 10 examples the effect of adding more diminishes. The difference in performance for the ``pull while spinning'' split can be explained by the high variance of these results.


\textbf{The effect of manner-similarity.} It turns out simply adding more adverbs to the training set does not suffice for strong systematic generalization. As detailed in the method section, the different gSCAN adverbs can be divided in three different groups. Adverbs in the same group require a similar manner of navigation. In this experiment (depicted in Figure \ref{fig:type}), we train a model on different subsets of the dataset containing 150 augmented adverbs in addition to the original four gSCAN adverbs (``All adverb types'' in Figure \ref{fig:type}). Sorted from most number of adverbs to least number of adverbs we have; a subset containing the ``while spinning'' and ``while zigzagging''-type adverbs but not the ``cautiously''-type (``No cautiously-type''), a disjoint subset with only the ``cautiously''-type adverbs, a subset containing just one adverb (picked to be highly similar to ``cautiously'', requiring an action sequence of ``turn\_right turn\_left turn\_left turn\_right'' before movement actions instead of ``turn\_left turn\_right turn\_right turn\_left''), and a subset without extra adverbs containing only the original four gSCAN adverbs (``No extra adverbs'' in Figure \ref{fig:type}). The results of this experiment  give a clearer picture than vocabulary size alone. It shows that the type of adverb is crucial for the performance. The model trained on the dataset without ``cautiously''-type adverbs does only slightly better on the few-shot ``cautiously" split than when trained without extra adverbs, but very well on the ``pull while spinning" split. In fact we see that taking out the adverbs that are similar to ``cautiously'' boosts performance significantly over using all 150 adverbs on the ``pull while spinning" split. This shows that the adverb similarity is important for knowledge transfer between adverbs and adding adverbs of another type can even hurt performance. When taking this experiment to the extreme by adding a single adverb that is similar to ``cautiously'' to the original adverbs (``one cautiously-type'' in Figure \ref{fig:type}), we observe that this almost explains all performance increase for the few-shot ``cautiously''-split. This experiment suggests that it is not the quantity, but the quality of data that we add that is important, and that the models may be relying on nearest-neighbor-style reasoning when generalizing to new adverbs.

\section{Conclusion}

In this work we investigate adverb-verb compositionality that has a transformative effect on action sequences required to perform a verb. This type of compositionality has seen little progress in the past. By applying two influential modeling principles, modularity and data augmentation, we set a new state-of-the-art on two grounded language understanding tests that evaluate this. Even though the method has an advantage over previous methods due to the augmented data, the results give insight into what can bring about progress on the set of tests that thus far have proven difficult. We find that naively adding substantial experience with different adverbs to the training set of a neural network is not sufficient for strong generalization, but when separating the architecture into different modules for high-level cognitive processes the model is able to generalize more systematically. Moreover, we find it is not the quantity of experience that matters, but the quality; adding additional adverbs to the training set that correspond to a different manner of navigation than the one that is tested barely results in transfer of the learned knowledge and can even hurt generalization, whereas adding a single adverb that is very similar to the tested adverb almost explains the entire performance increase. These lessons  may translate more broadly; to move towards truly systematic neural networks we may need to take into account the quality of experience a model sees, and the inductive biases it needs to be able to utilize this experience systematically. For future work, it would be an interesting challenge to eliminate the need for designing a structured augmentation method by hand.
